\documentclass[letterpaper, 10 pt, conference]{ieeeconf}  

\IEEEoverridecommandlockouts    

\usepackage{verbatim}
\usepackage{todonotes}
\usepackage{amsfonts}
\usepackage{amsmath}
\usepackage{hyperref}
\usepackage{soul}
\usepackage{xspace}
\usepackage{subcaption}
\usepackage[font=small]{caption}

\begin{document}

\title{\LARGE \bf Learning energy-efficient driving behaviors by imitating experts}

\author{Abdul Rahman Kreidieh$^{1,2}$, Zhe Fu$^{2}$, and Alexandre M. Bayen$^{2}$
\thanks{$^{1}$Corresponding author, email: aboudy@berkeley.edu}%
\thanks{$^{2}$University of California, Berkeley}%
}

\maketitle

\begin{abstract}
The rise of vehicle automation has generated significant interest in the potential role of future automated vehicles (AVs). In particular, in highly dense traffic settings, AVs are expected to serve as congestion-dampeners, mitigating the presence of instabilities that arise from various sources. However, in many applications, such maneuvers rely heavily on non-local sensing or coordination by interacting AVs, thereby rendering their adaptation to real-world settings a particularly difficult challenge. To address this challenge, this paper examines the role of imitation learning in bridging the gap between such control strategies and realistic limitations in communication and sensing. Treating one such controller as an ``expert", we demonstrate that imitation learning can succeed in deriving policies that, if adopted by 5\% of vehicles, may boost the energy-efficiency of networks with varying traffic conditions by 15\% using only local observations. Results and code are available online at \url{https://sites.google.com/view/il-traffic/home}.
\end{abstract}

\section{Introduction}

Autonomous driving systems have the potential to vastly improve the quality and efficiency of existing transportation systems. With fast reaction times and socially optimal behaviors, \emph{automated vehicles} (AVs) can improve the road capacity and traffic flow stability of existing networks~\cite{wadud2016help,talebpour2016influence,stern2018dissipation,wu2017emergent,kreidieh2018dissipating}, as well as reduce instances of vehicle collisions and similar incidents brought about by human errors~\cite{blanco2016automated}. The promise of such systems, however, is seldom witnessed in practice~\cite{gunter2020commercially,claybrook2018autonomous}, highlighting the challenge that lies ahead.

For the condition of addressing traffic flow instabilities in highway networks, longitudinal motion planning systems have been at the core of the conversation~\cite{zhou2019longitudinal,siri2021freeway}. Even in \emph{mixed-autonomy} settings, whereby only a subset of vehicles act as AVs, such systems have been demonstrated to provide significant improvements to both network throughput and energy-efficiency. In their seminal work, for instance, the authors of~\cite{stern2018dissipation} empirically demonstrate that even at low penetration rates, AVs driving at the equilibrium free-flow speed of a single-lane ring road can effectively stabilize traffic conditions for all vehicles involved, reducing system-level fuel consumption by $40$\% in the process. Such maneuvers, however, rely on prior knowledge of free-flow conditions, or access to non-local sensing and communication apparatuses from which to estimate them~\cite{seo2017traffic}, making these maneuvers difficult to adapt to arbitrary highway networks.

In this paper, we demonstrate that \emph{imitation learning} (IL) techniques can help alleviate the need for non-local traffic state estimates in expertly designed AV controllers. Looking to multi-lane highways in particular, we construct a controller inspired by the work of~\cite{stern2018dissipation} that dissipates the formation of stop-and-go traffic, but relies on downstream information to do so. Next, we apply IL to the controller and validate that such techniques, when properly deployed, can learn patterns in locally observable features that negate the need for such information, resulting in AV driving policies that operate as if they are knowledgeable of the state of downstream traffic. These findings suggest that techniques such as IL can serve an important role in adapting traffic smoothing controllers to limitations that arise in the real world.

The primary contributions of this paper are:

\begin{itemize}
    \item We design a controller (serving as the expert) that, through some degree of non-local sensing, dissipates the formation of stop-and-go waves within open highway networks.
    \item We identify key limitations in both the expert model and training procedure that limit the efficacy of IL and describe methods for addressing them. These include introducing temporal reasoning through the state representation and modifying the expert to allow it to recover from errors.
    \item Finally, we study the performance and robustness of the imitated policies, demonstrating that the policy succeeds at nearly perfectly matching the performance of the expert without the need for non-local state information.
\end{itemize}

The remainder of this paper is organized as follows. Section~\ref{sec:prelims} provides an overview of imitation learning and mixed-autonomy traffic. Section~\ref{sec:problem-setup} outlines the problem setup, expert, and IL approach. Section~\ref{sec:results} presents the findings and results of computational experiments over a wide variety of traffic states and compares the performance of the expert and imitated policies. Finally, Section~\ref{sec:discussion-future-work} provides concluding remarks and potential avenues for future work.

\section{Related Work} \label{sec:prelims}


\subsection{Traffic congestion and mixed-autonomy traffic} \label{sec:congestion-mixed-autonomy}

Traffic congestion is a state of traffic characterized by slower speeds, longer trip times, and increased instances of vehicular queuing. The presence of such phenomena degrades the quality and efficiency of existing networks, resulting in reduced fuel efficiency~\cite{treiber2008much} and increased incidents of road rage~\cite{hennessy1999traffic} and vehicle-to-vehicle collisions~\cite{marchesini2010relationship}. As a result, solutions for such phenomena have been of primary interest within the transportation community.

The present study focuses on a form of congestion common to typical highways. Within these settings, congestion appears as \emph{stop-and-go waves}, characterized by periodic, sudden, and at times sharp oscillations in driving speeds. Interestingly, these waves emerge even in the absence of external disturbances. For example, in simplified networks such as ring roads, small fluctuations arising from heterogeneities in human driving behaviors can produce higher amplitude oscillations by following vehicles as they break harder to avoid unsafe settings~\cite{sugiyama2008traffic}. This response, termed \emph{string instability}~\cite{swaroop1996string}, results in the formation of stop-and-go waves as oscillations propagate deeper through a platoon.

The presence of string instabilities in human driving highlights a potential benefit for upcoming automated driving systems. In particular, if properly implemented, AVs can reduce the effects of such instabilities amongst one another~\cite{lu2002acc}, and by influencing the behaviors of neighboring vehicles may succeed at mitigating congestion in \emph{mixed autonomy} settings, where only a subset of vehicles are automated. In~\cite{stern2018dissipation}, for instance, the authors demonstrate that stop-and-go waves in ring road settings can be dampened by operating as few as 5\% of vehicles at the effective \emph{equilibrium speed} of the network. The control strategy applied to the AVs, termed the \emph{Follower Stopper}, 
acts as a velocity controller, navigating the speeds of automated vehicles to a predefined desired speed $U$ (set to the equilibrium speed) while maintaining a safe gap with the vehicle ahead. Following this model, the command velocity $v^\text{cmd}$ of an AV is defined as:
\begin{equation}
\resizebox{.9\hsize}{!}{$
v^\text{cmd} = \begin{cases}
0 & \text{if } h_\alpha \leq \Delta x_1 \\
v \frac{\Delta x - \Delta x_1}{\Delta x_2 - \Delta x_1} & \text{if } \Delta x_1 \leq h_\alpha \leq \Delta x_2 \\
v + (U - v) \frac{\Delta x - \Delta x_2}{\Delta x_3 - \Delta x_2} & \text{if } \Delta x_2 \leq h_\alpha \leq \Delta x_3 \\
U & \text{otherwise}
\end{cases}
$}
\label{eq:fs}
\end{equation}
where $v = \text{min}(\text{max}(v_l,0),U)$ and $\Delta x_k$ is defined as:
\begin{equation}
\Delta x_k = \Delta x_k^0 + \frac{1}{2d_k}(\Delta v_{-})^2, \ \ k = 1, 2, 3
\end{equation}
where $\Delta v_- = \min (\Delta v, 0)$ is the negative arm of difference between the speed of the lead vehicle and the AV. This intermediary desired speed in then converted to approach acceleration or torque commands via some form of lower-level control. In this paper, we exploit this final controller when designing our congestion-mitigating expert.
%
The parameters used for this model are provided in Table~\ref{table:model-params}.

\begin{figure*}
    \centering
    \begin{subfigure}[b]{\textwidth}
        \includegraphics[width=\linewidth]{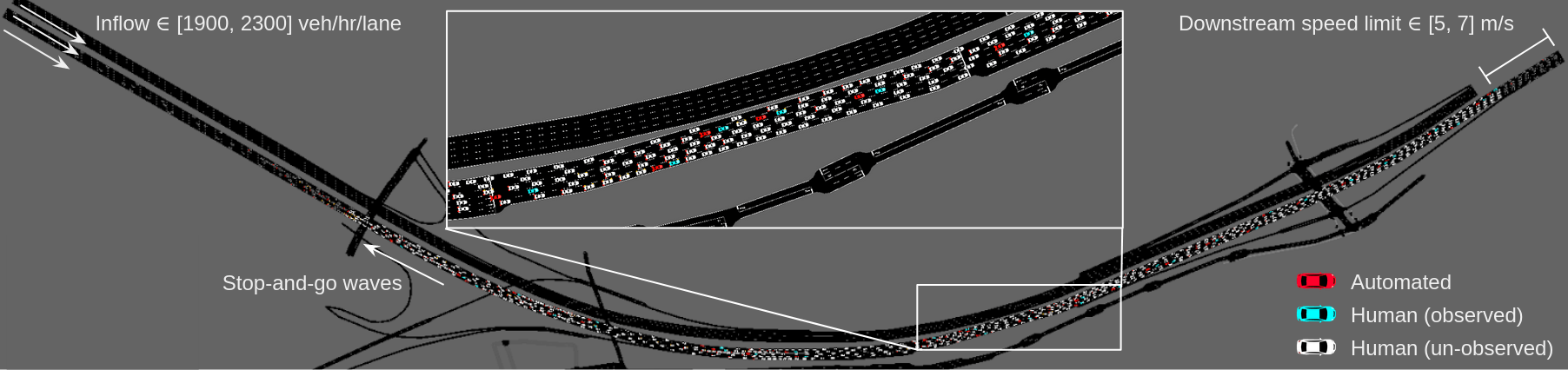}
        \caption{Network configuration}
        \label{fig:network}
    \end{subfigure}\\[5pt]
    \begin{subfigure}[b]{0.48\textwidth}
        \includegraphics[width=\linewidth]{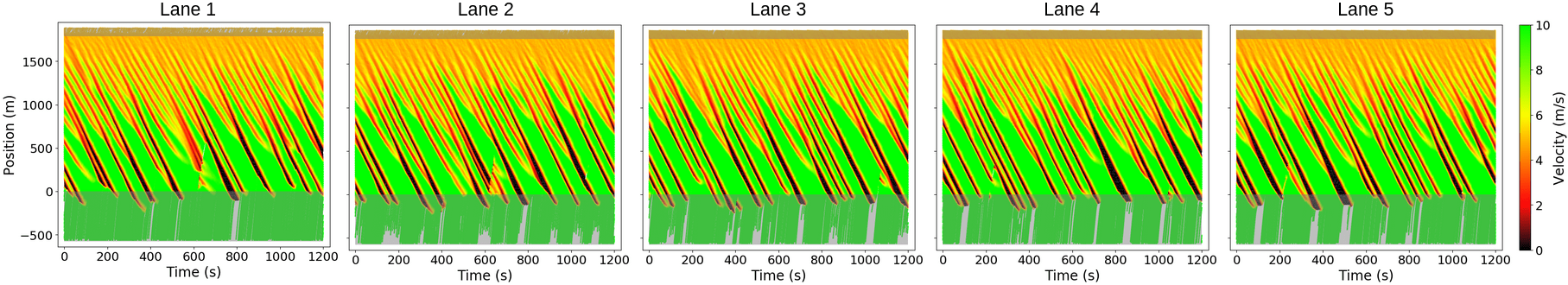}
        \caption{Human-driven baseline}
        \label{fig:human-performance}
    \end{subfigure}
    \hfill
    \begin{tikzpicture}
        \draw [dashed] (0,0) -- (0,2.2);
    \end{tikzpicture}
    \hfill
    \begin{subfigure}[b]{0.48\textwidth}
        \includegraphics[width=\linewidth]{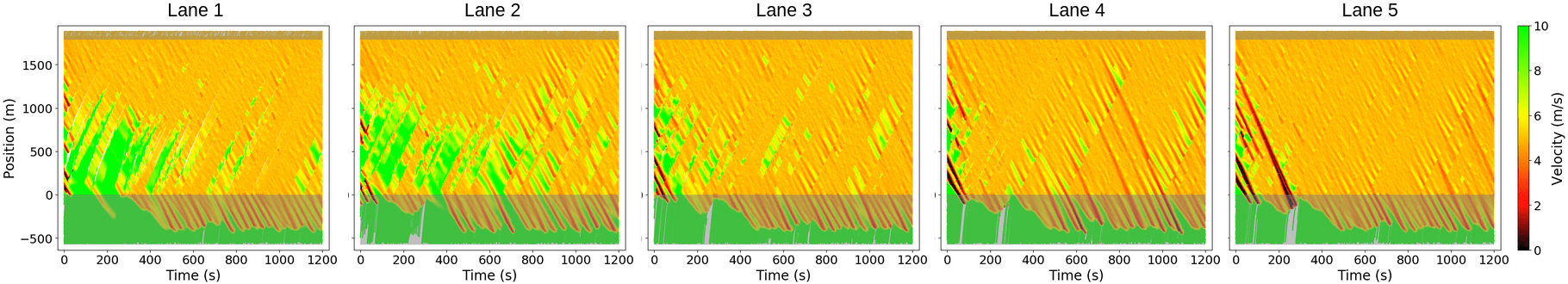}
        \caption{Expert controller performance}
        \label{fig:fs-performance}
    \end{subfigure}
    \caption{An illustration of the explored network. \textbf{Top:} We explore the application of congestion-mitigating strategies on a section of the I-210 network. \textbf{Bottom left:} In the absence of automation, restrictions to the outflow of vehicles from a slower-moving downstream edge result in the formation of stop-and-go waves. \textbf{Bottom right:} To mitigate congestion, we design an expert inspired by the Follower Stopper. This expert, which exploits downstream information from the network, can succeed in reducing the propagation of waves.}
    \label{fig:ant_gather_joint_opt}
\end{figure*}
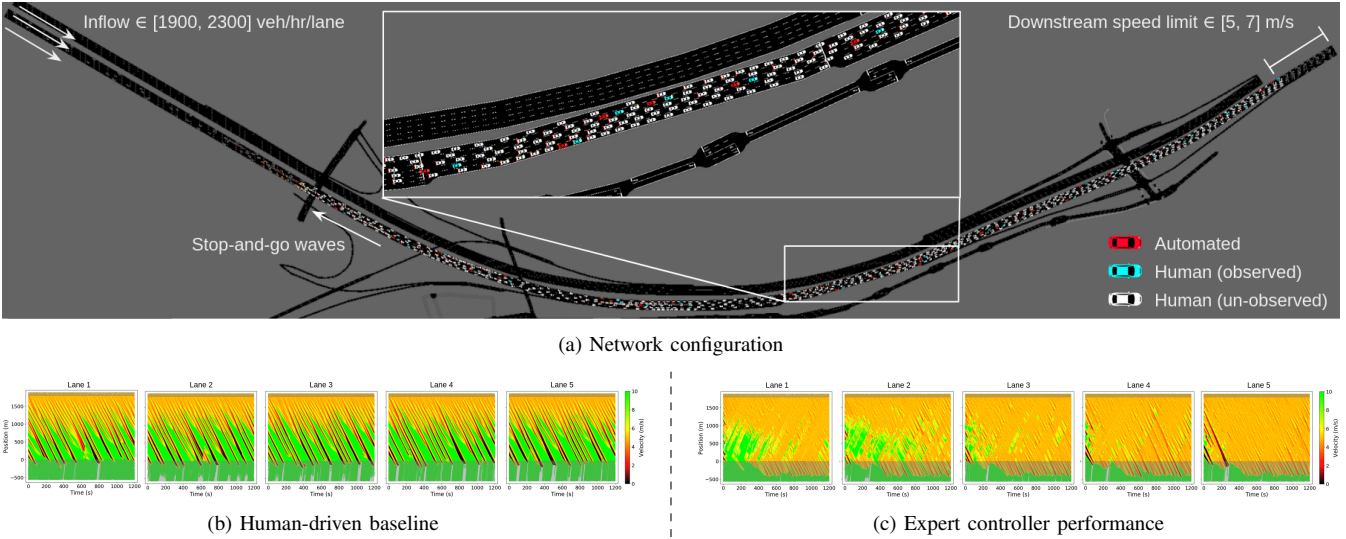

\begin{table}[]
\centering
\begin{scriptsize}
\begin{tabular}{|l|c|c|c|c|c|c|c|}
\hline
& \multicolumn{7}{c|}{Intelligent Driver Model (IDM)} \\ \hline
\textbf{Parameter} & $v_0$ & $T$ & $a$ & $b$ & $\delta$ & $s_0$ & $\epsilon$ \\ \hline
\textbf{Value} & $30$ & $1$ & $1.3$ & $2.0$ & $4$ & $2$ & $\mathcal{N}(0,0.3)$ \\ \hline
\end{tabular}\\[7pt]

\begin{tabular}{|l|c|c|c|c|c|c|c|}
\hline 
& \multicolumn{7}{c|}{Follower Stopper} \\ \hline
\textbf{Parameter} & $\Delta x_1^0$ & $\Delta x_2^0$ & $\Delta x_3^0$ & $d_1$ & $d_2$ & $d_3$ & $a_{\max}$ \\ \hline
\textbf{Value}     & $4.5$ & $5.25$ & $6.0$ & $1.5$ & $1.0$ & $0.5$ & $1$ \\ \hline
\end{tabular}
\end{scriptsize}
\caption{Model parameters}
\label{table:model-params}
\end{table}

\subsection{Imitation learning} \label{sec:prelim-imitation}

Imitation learning (IL) refers to a class of algorithms that attempt to develop a policy $\pi_\theta(s_t) : \mathcal{S} \to \mathcal{A}$, parametrized by $\theta$, that matches the behavior of an expert $\pi^*$ through demonstrations of the performance of the expert within an environment. Let $\mathcal{D} = \left\{ (s_i, a_i) \right\}_{i=1}^N$ denote a dataset of (state, action) demonstrations provided by an expert. In this setting, the objective of an IL algorithm is to solve the problem: $\theta := \text{argmin}_\theta \left[ \mathbb{E}_{\mathcal{D}} \left[ \mathcal{L}(\pi_\theta(s_i), a_i) \right] \right]$, where $\mathcal{L}(\cdot, \cdot)$ is a distance metric between the predicted and expert action. For this work, we use a mean-square error loss function.

Imitation learning, and its role in autonomous driving, has grown rapidly since its initial deployment via ALVINN~\cite{pomerleau1989alvinn}. Nowadays, IL methods in autonomous driving have become valuable tools for model compression. In~\cite{sun2018fast}, for instance, these methods are used to compress computationally complex AV control strategies based on model-predictive control (MPC) into faster neural network representations. More broadly, IL has been actively being explored as a means of reducing the complexity of ruled-based approaches for autonomous driving into relatively simple end-to-end models trained to replicate the behaviors of human drivers~\cite{bansal2018chauffeurnet,codevilla2018end}. These findings and similar ongoing work highlight the potential benefits and adaptability of IL within the domain of driving. Taking inspiration from the above studies, we explore the ability of IL to compress behaviors of AV models reliant on knowledge from non-local state information into policies that are more conducive for real-world settings.

\section{Experimental Setup} \label{sec:problem-setup}

In this section, we detail the considered problem and design an expert controller which, using non-local estimates of traffic, can dissipate the formation and propagation of congestion in the considered setting. We then present an IL paradigm through which the presented controller can be redefined to dissipate congestion in arbitrary traffic conditions using only locally perceptible state information.

\subsection{Network configuration} \label{sec:network}

We explore the problem of designing congestion-mitigating strategies for AVs in multi-lane highways. The network considered, see Figure~\ref{fig:network}, is a simulation of a $1$-mile section of the I-210 network in Los Angeles, California. This network has been the topic of considerable research in recent years, with various studies aiming to identify and reconstruct the source of congestion within it~\cite{gomes2004congested,dion2015connected}. This paper in particular builds on the work of~\cite{lee2021integrated}, which explores the role of mixed-autonomy traffic systems in improving the energy-efficiency of vehicles within I-210.

Within this network, congestion is generated via an imbalance between the inflow and outflow conditions. In particular, high inflow rates matching peak demand intervals are provided from the leftmost edge, and outflow rates are restricted via a reduced speed limit within the rightmost $100$~m. These restrictions serve to model various downstream effects, including from slow-moving or stopping bottlenecks and downstream arterial or merges networks. The overall imbalance increases the density of the network, which coupled with string-instabilities in human-driver dynamics\footnote{To model string instabilities within the network, we use the \emph{Intelligent Driver Model} (IDM)~\cite{treiber2000congested}. The specific model parameters are in Table~\ref{table:model-params}.} results in the onset of stop-and-go traffic. Figure~\ref{fig:human-performance} depicts this behavior in the fully human-driven setting.

\subsection{Dissipating stop-and-go waves via non-local sensing}

To dissipate the emergence of stop-and-go waves within this network in mixed-autonomy settings, we seek to adapt to the work of~\cite{stern2018dissipation} (see Section~\ref{sec:congestion-mixed-autonomy}) to multi-lane highways. To do so, we must first define a distribution of desired speeds that consistently dissipate the formation of waves within the network. For the problem explored within this paper, we take inspiration from ramp-metering systems for traffic signal control~\cite{papageorgiou1991alinea,papageorgiou2002freeway}. Through such systems, free-flow conditions in high-demand, throughput-restricted networks are sustained by bounding the inflow of vehicles below a network's breakdown capacity. Similarly, by driving near the outflow conditions of a given network, AVs can restrict the flow of vehicles upstream, thereby preventing the buildup of dense traffic and formation of subsequent stop-and-go waves. We accordingly choose to set the desired speed $U$ of AVs within the Follower Stopper to match the downstream speed of the network.


Figure~\ref{fig:fs-performance} depicts the performance of the aforementioned controller for an AV penetration rate of 5\%. Seen here, the AVs succeed at dissipating the formation of stop-and-go waves, resulting in more uniform driving speeds across the network. This is true for a range of downstream boundaries conditions where stop-and-go congestion occurs and results in significant reductions to energy consumption. This solution, however, requires knowledge of downstream information, which may be unavailable. In the following subsection, we explore methods for subverting this limitation via IL.

\subsection{Absolving the dependence on non-local states} \label{sec:method-learning}

In this paper, we aim to determine whether control strategies similar to the one presented previously may be derived from features observable to individual AVs. Specifically, for a given AV $\alpha$, we seek to design a controller $\pi_\theta(v_\alpha, h_\alpha, v_l)$ which maps actions by the expert Follower Stopper controller $a_\text{FS}(v_\alpha, h_\alpha, v_l, U)$ solely to the variables $v_\alpha$, $v_l$, $h_\alpha$, while abolishing the reliance on the non-local desired speed term $U$. To do so, we explore the use of the IL methods described in Section~\ref{sec:prelims} while treating the Follower Stopper as the expert. This process introduces several challenges that must be addressed. We describe the two most prominent challenges and introduce techniques to mitigate them in practice. 

\subsubsection{Challenge \#1: Unobservability of downstream state}

The first of these challenges accounts for the absence of non-local observations for the imitated policy. The use of purely local observations confounds the expert's behavior across varying downstream conditions, thereby making specialization to unique downstream events difficult. In response to this challenge, we hypothesize that local historical data can be exploited to estimate traffic flow properties further downstream. In particular, fluctuations in driving speeds from the perspective of an AV, or the act of approaching or moving away from a lead vehicle over time, may provide useful information on the effective equilibrium speed of a given network in relation to one's own. To validate this assumption, we introduce additional temporal information to the state space of the imitated policy. Specifically, we concatenate to the original observation of an AV the past $N$ states, collected at a sampling rate of $\Delta t$. For this problem, the following constants are chosen: $N=5$ and $\Delta t=2$.

\begin{figure}
\centering
\includegraphics[width=\linewidth]{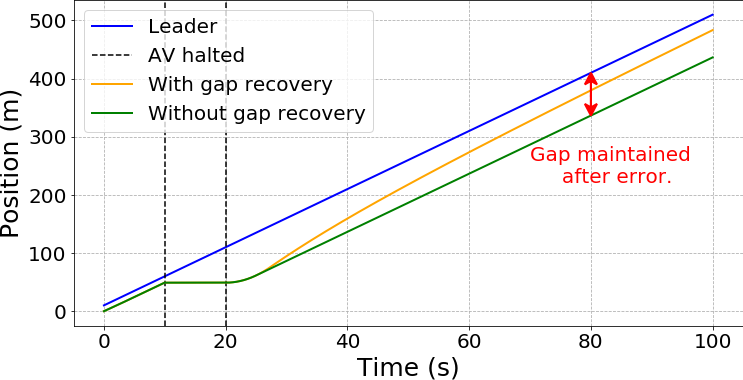}
\caption{Behavior of the Follower Stopper when following a leader with and without gap recovery. The original controller, in green, cannot address gaps after an error is introduced. Conversely, the modified controller, in orange, tends towards its prior trajectory.}
\label{fig:gap-recovery}
\end{figure}

\begin{figure*}
\centering
\begin{subfigure}[b]{0.9\textwidth}
    \begin{subfigure}[b]{\textwidth}
        \includegraphics[width=\linewidth]{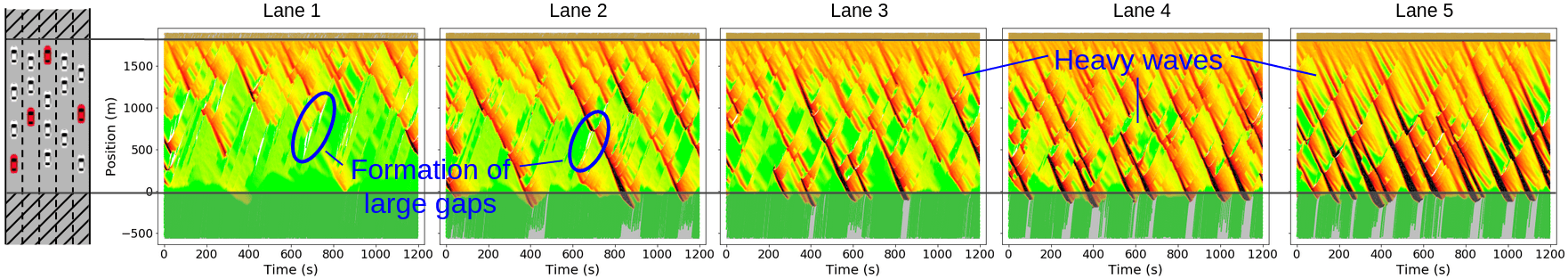}
        \caption{Without gap recovery}
    \end{subfigure}
    \\[5pt]
    \begin{subfigure}[b]{\textwidth}
        \includegraphics[width=\linewidth]{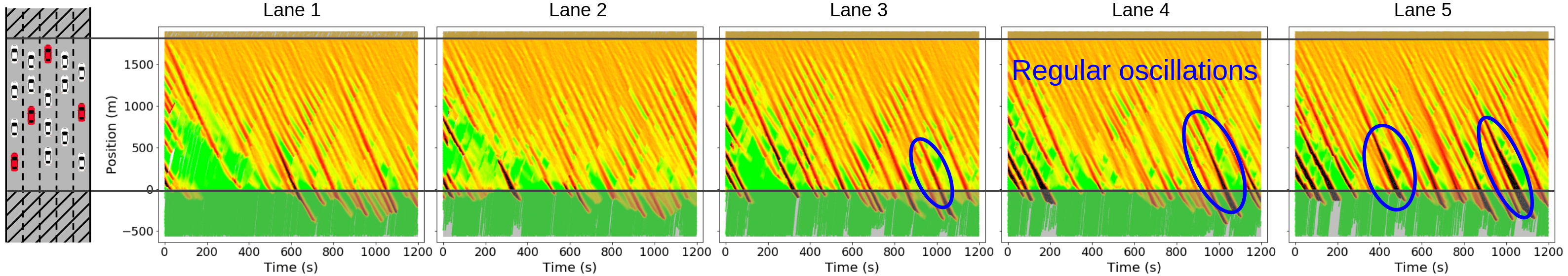}
        \caption{Without temporally extended observations}
    \end{subfigure}
    \\[5pt]
    \begin{subfigure}[b]{\textwidth}
        \includegraphics[width=\linewidth]{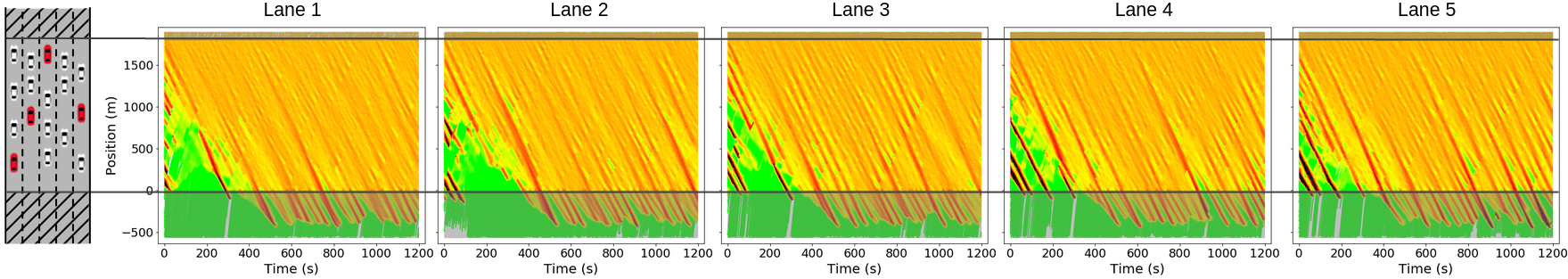}
        \caption{Final approach}
        \label{fig:final-approach}
    \end{subfigure}
\end{subfigure}
\hfill
\begin{subfigure}[b]{0.075\textwidth}
    \includegraphics[width=\linewidth]{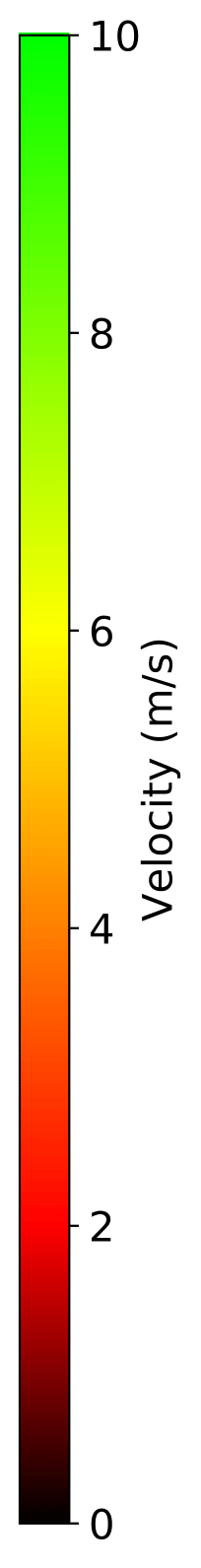}
\end{subfigure}
\caption{Spatio-temporal comparison of the imitated model for an inflow rate of $2100$ veh/hr/lane and downstream speed limit of $5$ m/s. \textbf{a)} When gap-recovery is not enforced within the expert, AVs learn to form large gaps with their leaders whenever possible, resulting in the emergence of more pronounced waves in adjacent lanes. \textbf{b)} When additional temporal knowledge is not provided, the imitated policy does performs well, but fails to dissipate waves as frequently as the expert. \textbf{c)} The final approach succeeds in identifying and driving at the effective speed of the network, and in doing so dissipates the vast majority of emerging stop-and-go waves in each individual lane.}
\label{fig:ablation}
\end{figure*}

\subsubsection{Challenge \#2: Recovering from errors}

The second of these limitations accounts for features within the expert that are non-conducive for IL. In many driving applications (e.g. lane-keeping~\cite{bojarski2016end}), IL methods rely on demonstrations of how to recovery from non-optimal or erroneous trajectories to produce safe and robust behaviors.
For longitudinal driving tasks explored here, errors emerge in the form of large gaps. In particular, at the early stages of training, the imitated policy often does not produce actions that maintain the speed desired by the expert, instead slowing down and allowing its leader to outpace it. While this is to be expected, the Follower Stopper does not inform policies how to recover from such errors, instead opting to maintain a constant speed despite the magnitude of the gap ahead. As a result, the presence of such errors and the absence of labels that correct for them introduces an incentive to maintain large headways, which may result in additional undesirable behaviors.

In order to address this challenge, we introduce an additive term to the final condition of the expert, thereby informing the AVs to reduce the gap when large values emerge. The modified Follower Stopper command speeds is:
\begin{equation}
\resizebox{.9\hsize}{!}{$
v^\text{cmd} = \begin{cases}
0 & \text{if } \Delta x \leq \Delta x_1 \\
v \frac{\Delta x - \Delta x_1}{\Delta x_2 - \Delta x_1} & \text{if } \Delta x_1 \leq \Delta x \leq \Delta x_2 \\
v + (U - v) \frac{\Delta x - \Delta x_2}{\Delta x_3 - \Delta x_2} & \text{if } \Delta x_2 \leq \Delta x \leq \Delta x_3 \\
U \textcolor{red}{+\ c (\Delta x - \Delta x_3)^2}  & \text{otherwise}
\end{cases}
$}
\end{equation}
where $c$ is a constant term that toggles the rate at which recovery occurs, and is set to $0.001$ in future experiments.

Figure~\ref{fig:gap-recovery} depicts the performance of the original and modified Follower Stopper in the presence of gap-inducing errors. The benefit of this modification becomes evident under this setting, with the modified controller clearing the gap and the original failing to do so.

\subsection{Simulation and training procedure} \label{sec:setup-simulations}

Experiments are conducted in Flow~\cite{wu2017flow}, an open-source framework designed to enable the integration of machine learning tools with microscopic simulations of traffic. Within Flow, simulations of the I-210 are executed in SUMO~\cite{krajzewicz2012recent} with step sizes of $0.4$~sec/step and are warmed-started for $3600$~sec to allow for the onset of congestion before being run within the training procedure an additional $600$~sec. Each simulation is initialized with a random set of boundary conditions, sampled from a distribution of [$1900$, $2300$] veh/hr/lane in increments of $50$ for the inflow conditions, and [$5$, $7$] m/s in increments of $1$ for the downstream speed limit. Once the warmup period is finished, $5$\% of vehicles are replaced with AVs whose actions are sampled either from the expert or policy to mimic a similar penetration rate.

For all experiments in this paper, we use DAgger~\cite{ross2011reduction} for imitating the desired behavior by the expert. We initialize the expert dataset with $30\text{k}$ samples extracted from $20$ rollouts of the expert for varying inflow rate and downstream speed limit conditions and, after each training epoch, aggregate the dataset with $7.5\text{k}$ new samples from states visited by the imitated policy. This is repeated for $100$ training epoch for a total of $750\text{k}$ samples.
Finally, 
a Multi-Layer Perceptron is used with hidden layers $(32, 32, 32)$ and a ReLU non-linearity. This policy is updated using Adam~\cite{kingma2014adam} with a learning rate of $0.001$ and batch size of $128$, and dropout~\cite{srivastava2014dropout} is employed to allow for increased robustness. 

\section{Numerical Results} \label{sec:results}

In this section, we present numerical results for the expert and imitated policy presented in the previous section. Through these results, we aim to answer the following: 

\begin{enumerate}
    \item Does the imitated policy succeed in improving the traffic stability and energy-efficiency of vehicles within the network for different traffic conditions?
    \item Is the introduction of temporal reasoning and gap recovery to the imitated and expert models, respectively, needed for proper training to occur?
    \item Is the imitated policy robust to unseen variations in network and human-driver behaviors?
\end{enumerate}

Imitation learning runs for the various experimental setups are executed over $5$ seeds, and the training performance is averaged and reported across all seeds to account for stochasticity between simulations and policy initialization.

\subsection{Performance comparison}

Figure~\ref{fig:final-approach} depicts the spatio-temporal performance of the imitated policy for a unique traffic state. As we can see, the learned behavior largely succeeds in mitigating the frequency and intensity of stop-and-go waves that arise in the fully human-driven setting (see Figure~\ref{fig:human-performance}), resulting in more uniform driving speeds throughout the network. Moreover, this behavior closely matches that by the expert in Figure~\ref{fig:fs-performance}, suggesting that the imitation procedure was successful.

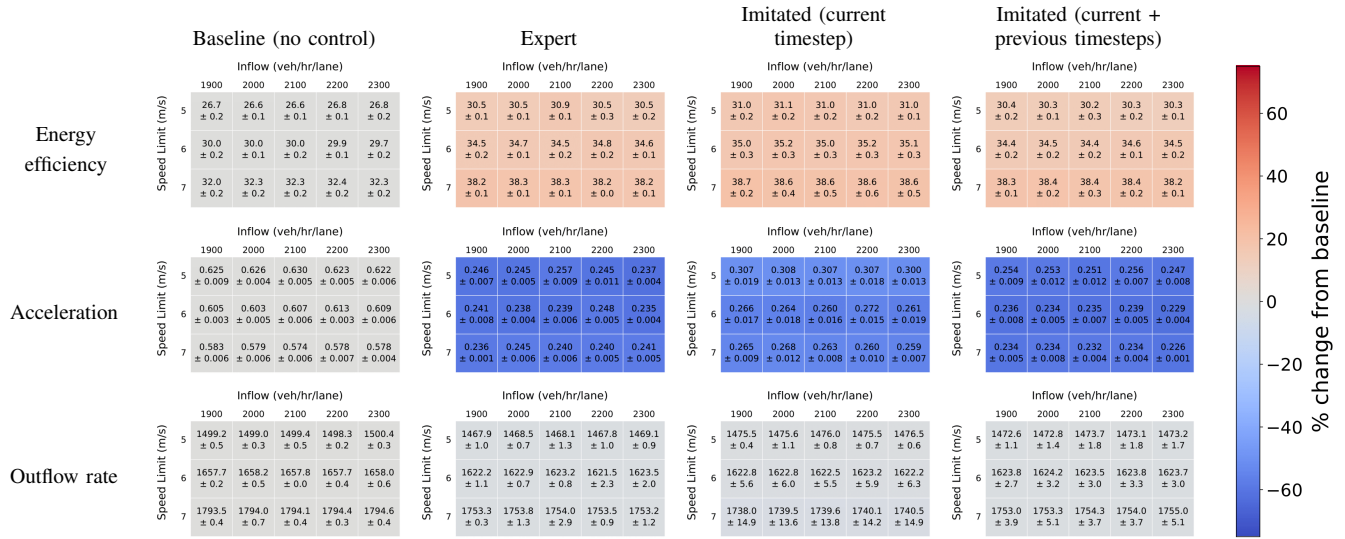
\begin{figure*}
\centering
\begin{subfigure}[b]{0.9\textwidth}
    \begin{subfigure}[b]{0.12\textwidth}
        \textcolor{white}{.}
    \end{subfigure}
    \hfill
    \begin{subfigure}[b]{0.21\textwidth}
        \begin{center}
            \footnotesize{Baseline (no control)}
        \end{center}
    \end{subfigure}
    \hfill
    \begin{subfigure}[b]{0.21\textwidth}
        \begin{center}
            \footnotesize{Expert}
        \end{center}
    \end{subfigure}
    \hfill
    \begin{subfigure}[b]{0.21\textwidth}
        \begin{center}
            \footnotesize{Imitated (current \\timestep)}
        \end{center}
    \end{subfigure}
    \hfill
    \begin{subfigure}[b]{0.21\textwidth}
        \begin{center}
            \footnotesize{Imitated (current + \\previous timesteps)}
        \end{center}
    \end{subfigure}
    \\[5pt]
    \begin{subfigure}[b]{0.12\textwidth}
        \begin{tikzpicture}
            \node at (-0.3,-0.5) {\textcolor{white}{.}};
            \node at (0.4,0.4) {\footnotesize{Energy}};
            \node at (0.4,0.0) {\footnotesize{efficiency}};
        \end{tikzpicture}
    \end{subfigure}
    \hfill
    \begin{subfigure}[b]{0.21\textwidth}
        \includegraphics[width=\linewidth]{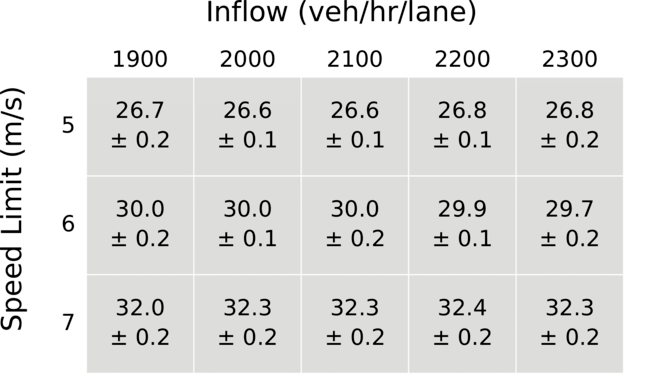}
    \end{subfigure}
    \hfill
    \begin{subfigure}[b]{0.21\textwidth}
        \includegraphics[width=\linewidth]{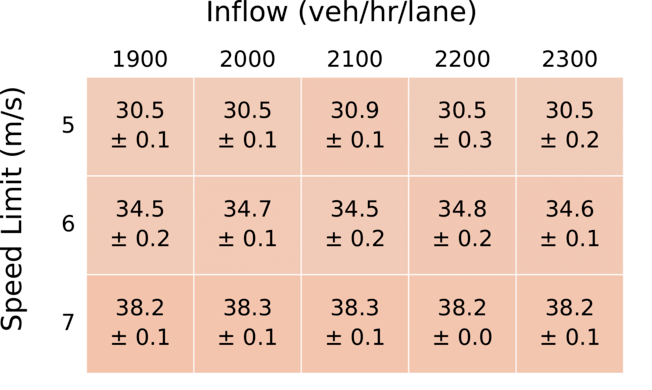}
    \end{subfigure}
    \hfill
    \begin{subfigure}[b]{0.21\textwidth}
        \includegraphics[width=\linewidth]{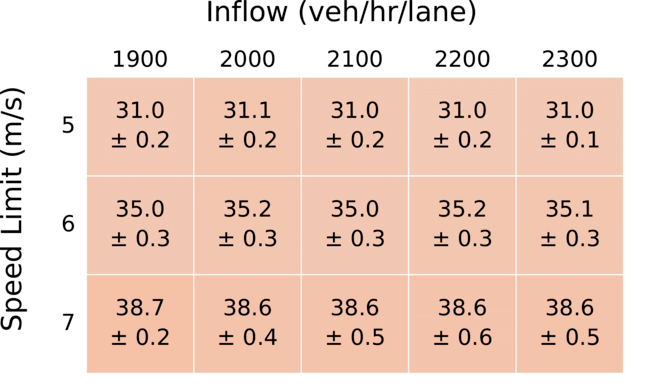}
    \end{subfigure}
    \hfill
    \begin{subfigure}[b]{0.21\textwidth}
        \includegraphics[width=\linewidth]{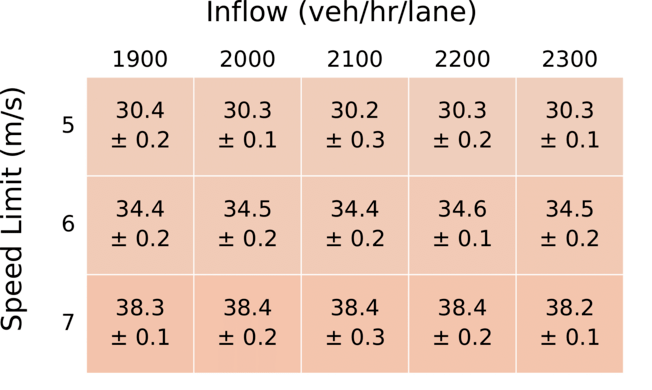}
    \end{subfigure}
    \\[5pt]
    \begin{subfigure}[b]{0.12\textwidth}
        \begin{tikzpicture}
            \node at (0,-0.75) {\textcolor{white}{.}};
            \node at (0,0) {\footnotesize{Acceleration}};
        \end{tikzpicture}
    \end{subfigure}
    \hfill
    \begin{subfigure}[b]{0.21\textwidth}
        \includegraphics[width=\linewidth]{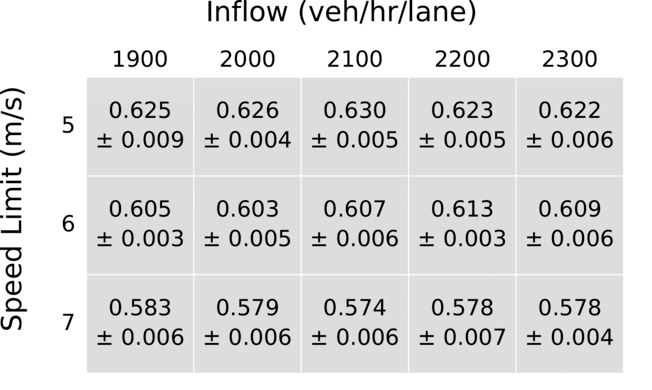}
    \end{subfigure}
    \hfill
    \begin{subfigure}[b]{0.21\textwidth}
        \includegraphics[width=\linewidth]{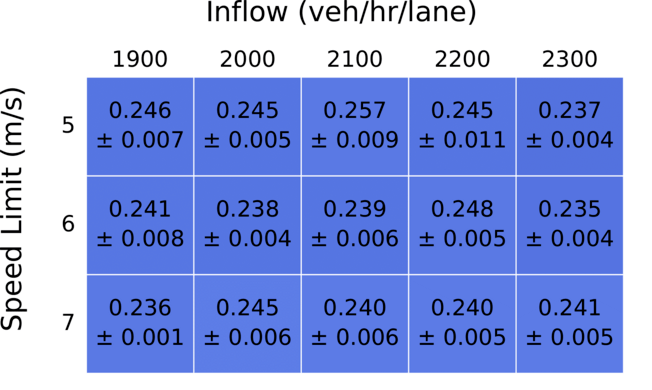}
    \end{subfigure}
    \hfill
    \begin{subfigure}[b]{0.21\textwidth}
        \includegraphics[width=\linewidth]{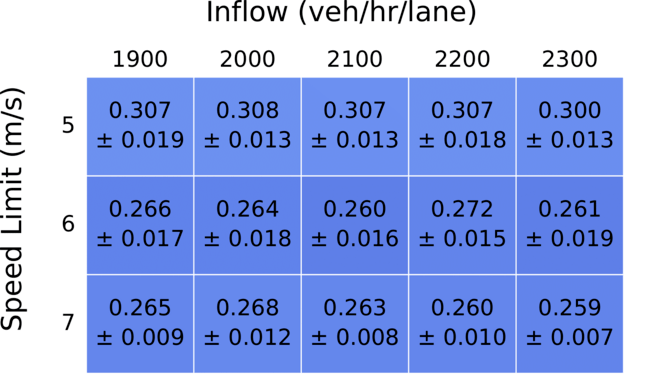}
    \end{subfigure}
    \hfill
    \begin{subfigure}[b]{0.21\textwidth}
        \includegraphics[width=\linewidth]{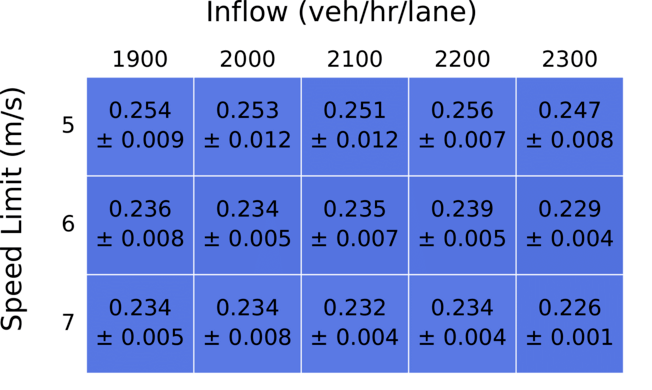}
    \end{subfigure}
    \\[5pt]
    \begin{subfigure}[b]{0.12\textwidth}
        \begin{tikzpicture}
            \node at (-0.5,-0.75) {\textcolor{white}{.}};
            \node at (0.15,0) {\footnotesize{Outflow rate}};
        \end{tikzpicture}
    \end{subfigure}
    \hfill
    \begin{subfigure}[b]{0.21\textwidth}
        \includegraphics[width=\linewidth]{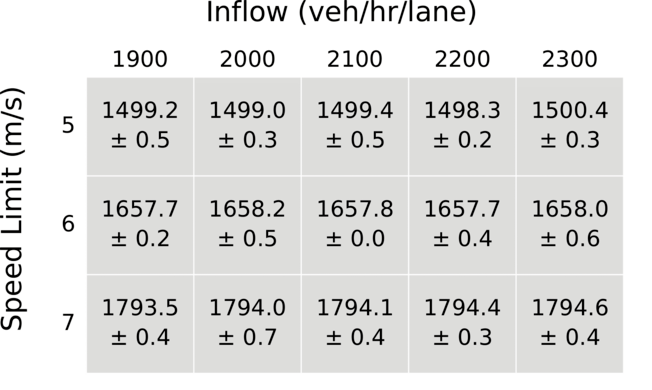}
    \end{subfigure}
    \hfill
    \begin{subfigure}[b]{0.21\textwidth}
        \includegraphics[width=\linewidth]{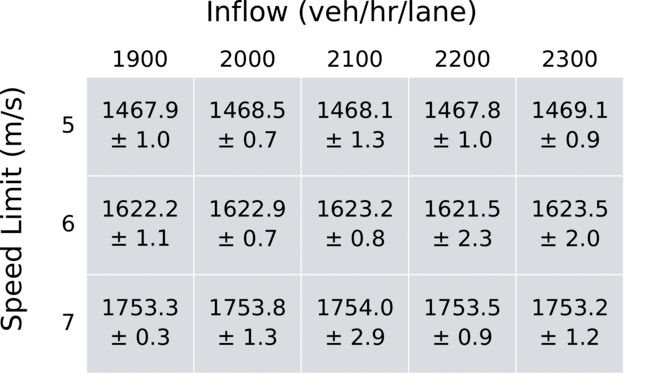}
    \end{subfigure}
    \hfill
    \begin{subfigure}[b]{0.21\textwidth}
        \includegraphics[width=\linewidth]{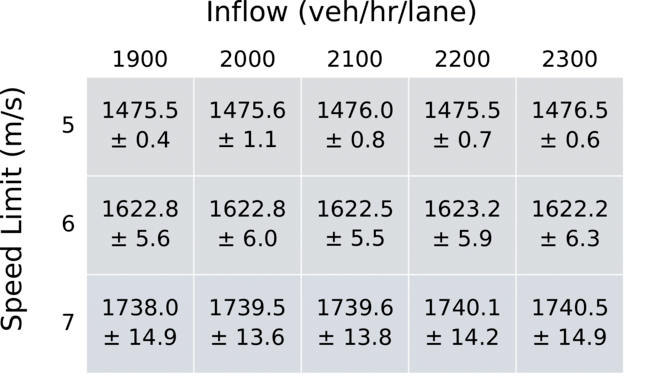}
    \end{subfigure}
    \hfill
    \begin{subfigure}[b]{0.21\textwidth}
        \includegraphics[width=\linewidth]{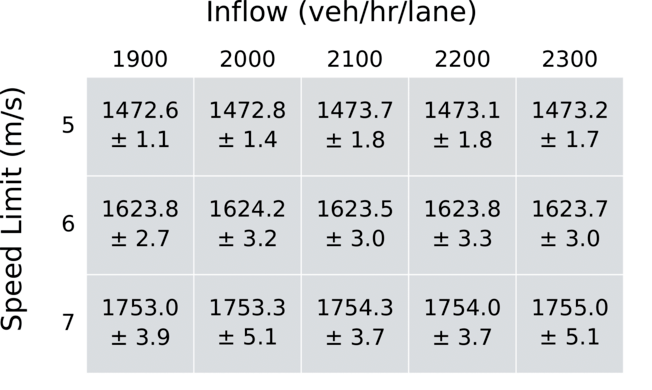}
    \end{subfigure}
\end{subfigure}
\hfill
\begin{subfigure}[b]{0.081\textwidth}
    \includegraphics[width=\linewidth]{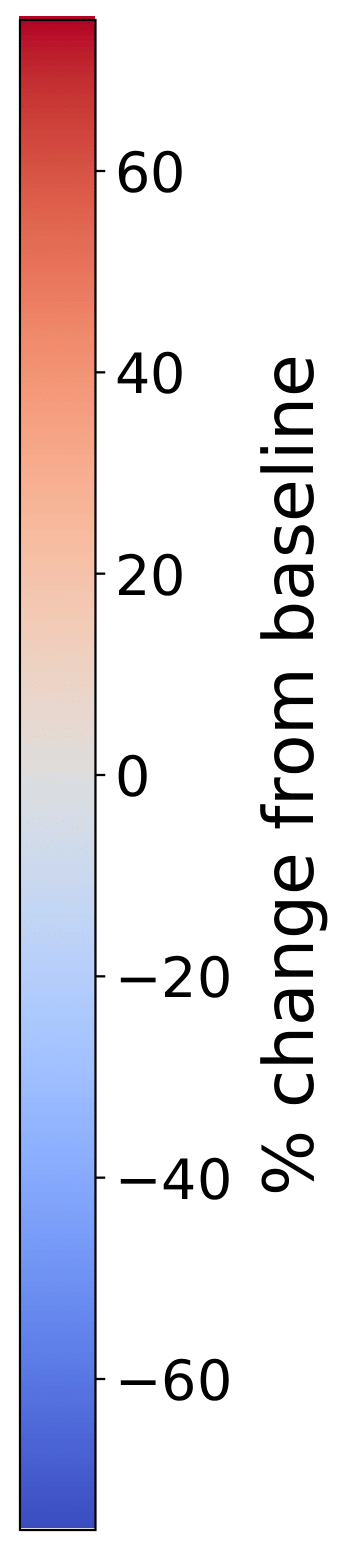}
\end{subfigure}
\caption{Performance of different models for different traffic conditions. Both the expert model (Follower Stopper) and imitated policies provide significant improvements to both energy-efficiency and the frequency and magnitude of accelerations without significantly degrading network throughput. This is true for all sampled inflow / downstream speed limit pairs.}
\label{fig:performance}
\end{figure*}

The performance of the imitated policy in comparison to the expert is further elucidated in Figure~\ref{fig:performance}. Within this figure, we compare the ability for both the expert and imitated policy to improve the energy-efficiency of the studied network based on two metrics: 1) a model of energy consumption, in miles per gallon, fitted to Toyota RAV4 emission data~\cite{lee2021integrated}, and 2) the magnitude of accelerations by both the human and automated vehicles within the network. We also compare the throughput of the network (in veh/hr/lane) to ensure any improvements to the energy-efficiency of the network do not emerge at the cost of mobility.
As we can see, the energy-efficiency of both the expert and imitated policy match one another very closely, producing values of $34.5$ mpg and $34.4$ mpg respectively in terms of energy consumption (a $15$\% improvement over the baseline value of $29.6$ mpg), and values of $0.242$ m/s$^2$ and $0.240$ m/s$^2$ respectively in terms of fluctuations in speed (a $60$\% reduction over the baseline value of $0.604$ m/s$^2$). The two also only experience very minor reductions in network throughput of around $2$\%, suggesting that the improvements to energy do not come at the cost of mobility. Remarkably, from the perspective of the imitated policy, this behavior emerges in the absence of non-local (downstream) knowledge of the state of traffic, demonstrating the potential for imitation learning to augment and even improve the quality of existing AV control strategies.

\subsection{Ablation studies} \label{sec:ablation}

Next, we explore the effect of the various augmentations presented in Section~\ref{sec:method-learning} on the performance of the resultant policy. Figure~\ref{fig:ablation} depicts the learned behaviors when a number of the proposed methods are removed from the training procedure. In the absence of gap recovery during the training procedure, the imitated policy produces undesirable large gaps in several lanes as marked by the red circles. These gaps, coupled with the lane-change dynamics of human vehicles, produce distributions of highly dense traffic in adjacent lanes that enable the further propagation of congestion, thereby negating the desired behaviors of the AVs as congestion mitigators. Alternatively, when gap recovery is enforced through the expert model but only the current time step is perceived through the input observation space, gaps do disappear and  waves are further dissolved, however, the regularity at which waves emerge is greater than is experienced when additional temporal information is provided to the imitated model.

The performance gap when temporally extended state information is not provided is further highlighted in Figure~\ref{fig:performance}. As seen in this figure, the use of current time-step information does perform well in terms of the studied energy-consumption model and network throughput when compared to the expert model. The disparity, however, is evident in terms of fluctuations in vehicle speeds, producing average acceleration/deceleration values of $0.278$ m/s$^2$, $15$\% greater than the expert model and imitated model with temporally extended information. This
concurs with the increased emergence of waves in Figure~\ref{fig:ablation}, and in the presence of vehicles with different energy emission properties may result in a degradation in the energy-efficiency of the network in comparison to more fine-tuned models. These findings reinforce the notion that temporal reasoning is needed to properly imitate models that utilize non-local state information. 

\subsection{Robustness to unseen phenomena}

Finally, we explore the robustness of imitated behaviors to unseen variations to the task at training time. In particular, we explore two types of variations. For one, we alter the penetration rate of AVs to determine whether the changes to the dynamics associated with such values negatively degrade the performance of the policy. Next, we look to variations in the responsiveness and aggressiveness of the car-following and lane-change models, respectively. Specifically, for the car-following behaviors, we alternate the maximum acceleration, or $a$ term, which controls the human's ability to respond to existing waves. Conversely, for the lane-change model, we modify the lane-change frequency parameter of the model, with larger values of this parameter resulting in more aggressive lane changes by the human drivers.

\begin{figure*}
\centering
\begin{subfigure}[b]{0.9\textwidth}
    \begin{subfigure}[b]{0.24\textwidth}
        \begin{center}
            \footnotesize{Baseline (no control)}
        \end{center}
    \end{subfigure}
    \hfill
    \begin{subfigure}[b]{0.24\textwidth}
        \begin{center}
            \footnotesize{Expert}
        \end{center}
    \end{subfigure}
    \hfill
    \begin{subfigure}[b]{0.24\textwidth}
        \begin{center}
            \footnotesize{Imitated (current \\timestep)}
        \end{center}
    \end{subfigure}
    \hfill
    \begin{subfigure}[b]{0.24\textwidth}
        \begin{center}
            \footnotesize{Imitated (current + \\previous timesteps)}
        \end{center}
    \end{subfigure}
    \\[5pt]
    \begin{subfigure}[b]{0.24\textwidth}
        \textcolor{white}{.}
    \end{subfigure}
    \hfill
    \begin{subfigure}[b]{0.24\textwidth}
        \includegraphics[width=\linewidth]{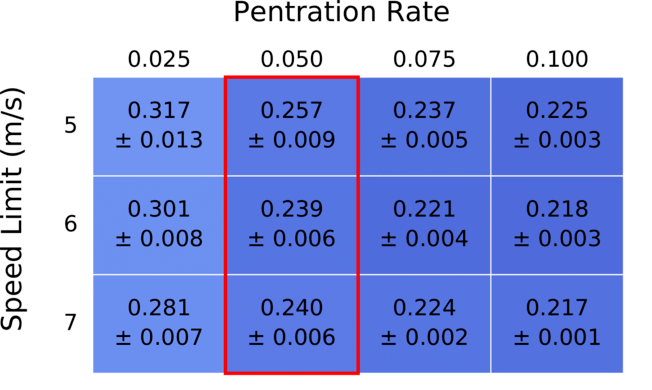}
    \end{subfigure}
    \hfill
    \begin{subfigure}[b]{0.24\textwidth}
        \includegraphics[width=\linewidth]{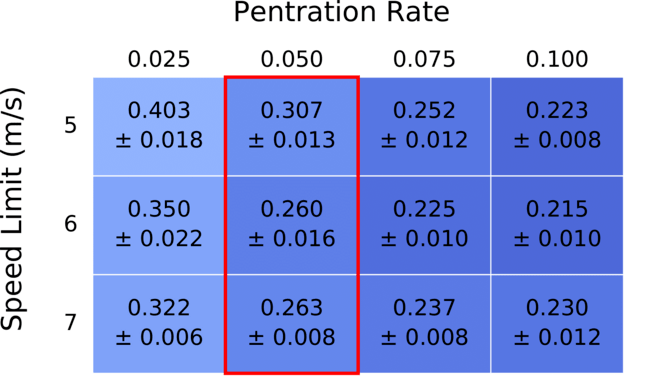}
    \end{subfigure}
    \hfill
    \begin{subfigure}[b]{0.24\textwidth}
        \includegraphics[width=\linewidth]{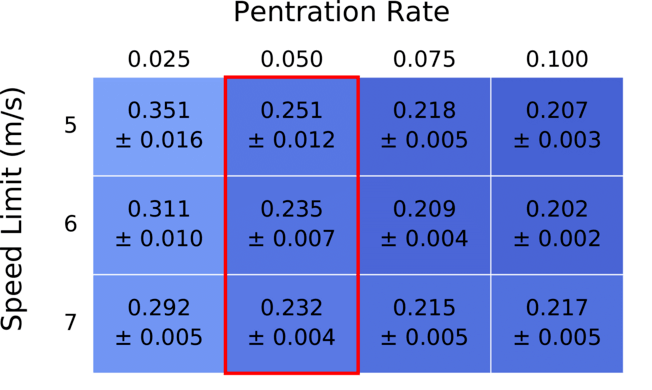}
    \end{subfigure}
    \\[-5pt]
    \begin{tikzpicture}
        \draw [dashed] (0,0) -- (\linewidth,0);
    \end{tikzpicture}
    \\[5pt]
    \begin{subfigure}[b]{0.24\textwidth}
        \includegraphics[width=\linewidth]{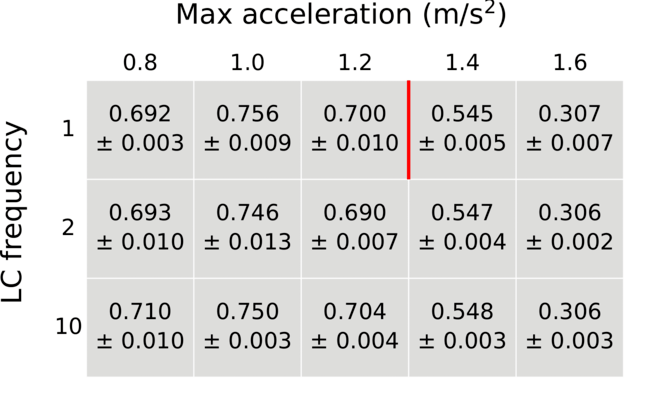}
    \end{subfigure}
    \hfill
    \begin{subfigure}[b]{0.24\textwidth}
        \includegraphics[width=\linewidth]{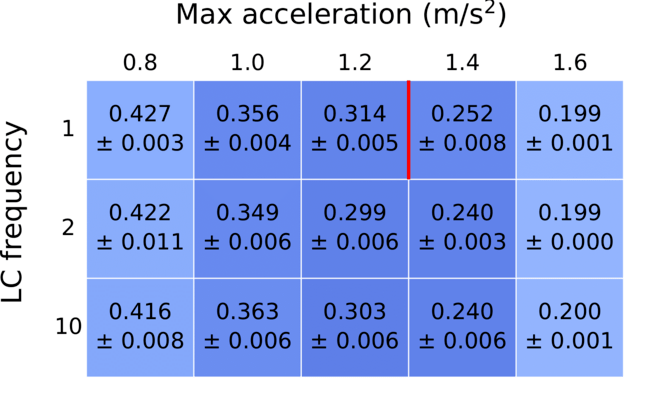}
    \end{subfigure}
    \hfill
    \begin{subfigure}[b]{0.24\textwidth}
        \includegraphics[width=\linewidth]{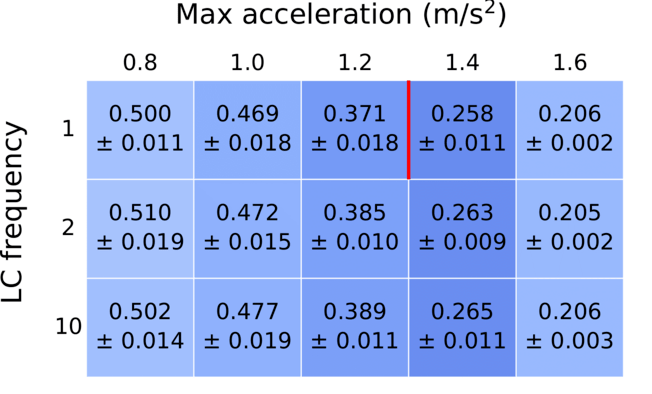}
    \end{subfigure}
    \hfill
    \begin{subfigure}[b]{0.24\textwidth}
        \includegraphics[width=\linewidth]{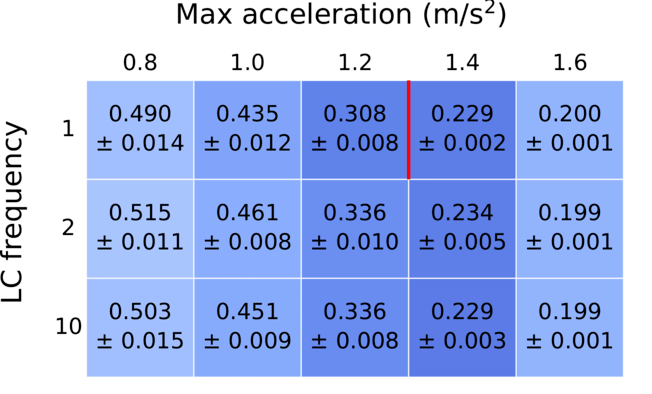}
    \end{subfigure}
\end{subfigure}
\hfill
\begin{subfigure}[b]{0.072\textwidth}
    \includegraphics[width=\linewidth]{figures/imitated/robustness/colorbar.png}
\end{subfigure}
\caption{Generalizability of the learned policy to different AV penetration rates (top) and human-driver conditions (bottom). The original training regime is bordered in red. We find that the imitated policy generalizes relatively well, continuing to outperform the baseline in all settings. The policy imitated with additional temporal information also succeeds in generally better outside its training scheme.}
\label{fig:generalizability}
\end{figure*}

Figure~\ref{fig:generalizability} depicts the performance of the baseline, expert, and imitated policies for the aforementioned variations in simulation parameters. For the variations in AV penetration rate, we explore the influence of varying downstream speed limits as well but set the inflow rate to a fixed value of $2100$ veh/hr/lane for each run.
Moreover, for the variations in human-driver model parameters, we restrict the analysis to a single pair of boundary conditions of $2100$ veh/hr/lane and $5$ m/s. As we can see from this figure, the imitated policies generalize well to different penetration rates, reducing slightly for low penetration rates of $2.5$\%, but outperforming the expert at higher penetration rates. For modifications to the human-driver models, a performance gap does begin to emerge between the expert and imitated policy for large changes to the parameters; however, the imitated policy does continue to outperform the human-driven baseline for all regions in which the expert outperforms as well.
The gap is also reduced once temporally extended state information is introduced, further highlighting the benefit of this approach.

\section{Conclusion} \label{sec:discussion-future-work}

This paper explores the problem of designing congestion-mitigating control strategies for automated vehicles in mixed-autonomy highways. For one, it demonstrates that such strategies can be achieved by aligning the speeds of a number of vehicles with the effective speed of downstream traffic, thereby preventing the onset of dense traffic. Next, through imitation learning it demonstrates that such strategies can be mapped to observations that are locally perceivable to individual vehicles, producing a purely decentralized controller that does not require feedback from non-local sources. This final behavior is shown to improve the energy-efficiency of the explored network for values as high as $15$\% while effectively eliminating the presence of most waves.

As a topic of ongoing work, we aim to determine whether similar control design strategies can be applied to real-world driving settings. In particular, we aim to validate the performance of this control strategy on trajectories of human driving provided by the real world. 

\section*{Acknowledgments}
This material is based upon work supported by the U.S. Department of Energy’s Office of Energy Efficiency and Renewable Energy (EERE) award number CID DE-EE0008872. The views expressed herein do not necessarily represent the views of the U.S. Department of Energy or the United States Government.

\bibliographystyle{IEEEtran}
\bibliography{main}

\end{document}